\documentclass[conference]{IEEEtran}
\IEEEoverridecommandlockouts
\usepackage{cite}
\usepackage{amsmath,amssymb,amsfonts}
\usepackage{algorithmic}
\usepackage{graphicx}
\usepackage[inkscapelatex=false]{svg}
\usepackage{textcomp}
\usepackage{dblfloatfix}
\usepackage{url}
\usepackage[caption=false,font=footnotesize]{subfig}
\usepackage{xcolor}
\def\BibTeX{{\rm B\kern-.05em{\sc i\kern-.025em b}\kern-.08em
    T\kern-.1667em\lower.7ex\hbox{E}\kern-.125emX}}
\begin{document}

\title{Large Language Model-assisted Autonomous Vehicle Recovery from Immobilization\\}

\author{\IEEEauthorblockN{1\textsuperscript{st} Zhipeng Bao}
\IEEEauthorblockA{\textit{School of Environmental, Civil, Agricultural, and Mechanical Engineering} \\
\textit{University of Georgia}\\
Athens, USA\\
zb28097@uga.edu} \\

\IEEEauthorblockN{2\textsuperscript{nd} Qianwen Li*}
\IEEEauthorblockA{\textit{School of Environmental, Civil, Agricultural, and Mechanical Engineering} \\
\textit{University of Georgia}\\
Athens, USA\\
cami.li@uga.edu}
}

\maketitle

\begin{abstract}
Despite significant advancements in recent decades, autonomous vehicles (AVs) continue to face challenges in navigating certain traffic situations that human drivers handle with ease. In such cases, AVs may become immobilized, leading to disruptions in overall traffic flow. Existing recovery solutions, including remote intervention, which is costly and inefficient, and manual takeover, which excludes non-drivers and limits accessibility, remain inadequate. This paper introduces StuckSolver, a novel Large Language Model (LLM) driven recovery framework that enables AVs to resolve immobilization scenarios through self-reasoning and passenger-guided decision making. StuckSolver is designed as a plug-in add-on module that operates on top of the AV’s existing perception, planning, and control stack without requiring modifications to its internal architecture. It interfaces directly with standard sensor data streams to detect immobilization states, interpret the environmental context, and generate high-level recovery commands that can be executed by the AV’s native planner. StuckSolver is integrated with CARLA’s rule-based Behavior Agent and evaluated using the Bench2Drive benchmark. Experimental results show that StuckSolver achieves driving scores and success rates on par with state-of-the-art end-to-end approaches through self-reasoning alone, and exhibits additional improvements when passenger guidance is incorporated. This confirms that StuckSolver provides robust operational performance within a lightweight and interpretable framework.\end{abstract}

\begin{IEEEkeywords}
AI agent, long-tail scenario, Bench2Drive, closed-loop simulation
\end{IEEEkeywords}

\section{Introduction}
After years of development, AVs have progressed beyond controlled environments such as airports and ports. Today, they are increasingly navigating more complex urban environments. Companies like Baidu and Waymo have deployed large fleets of robotaxis in cities such as Beijing, Wuhan, Phoenix, San Francisco, and Los Angeles, providing public services. For instance, Waymo’s robotaxis offer approximately 150,000 paid rides weekly, collectively traveling over one million miles ~\cite{Waymo2024}. This expansion marks a significant step toward integrating AVs into everyday urban mobility. 

\begin{figure}[t]
  \centering
   \includegraphics[width=8.5cm, height=9.5cm]{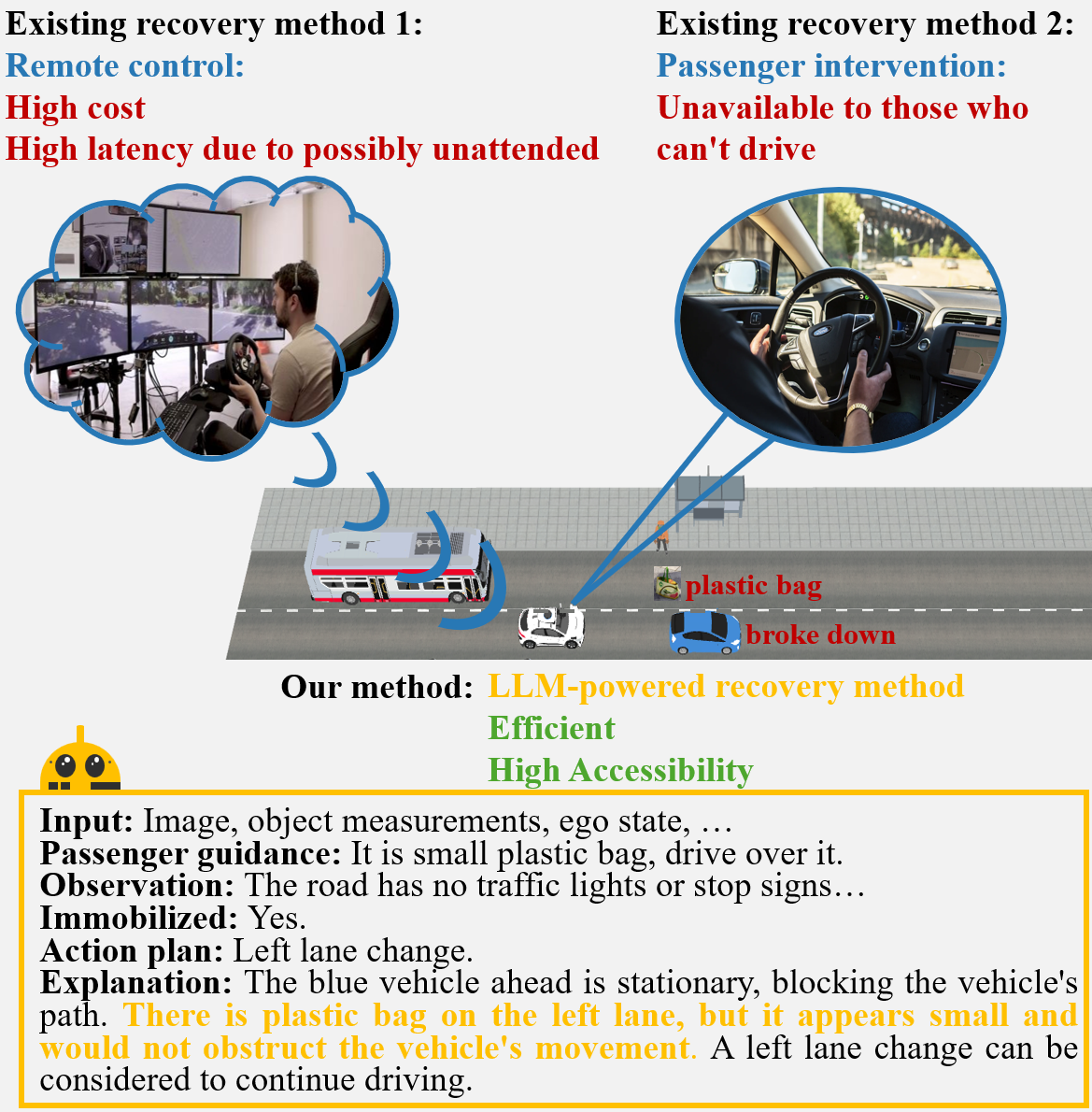}

   \caption{Illustration of AV immobilization in traffic. The white AV, carrying non-driving passengers, travels in the right lane. When the leading blue vehicle becomes disabled and the left lane is obstructed by a paper bag, the AV’s re-planning module fails to generate an alternative trajectory due to the perceived obstruction. Although the bag is a benign and traversable object, the AV misclassifies it as a non-traversable obstacle, resulting in a complete stop and prolonged immobilization.}
   \label{fig:1}
\end{figure}

\begin{figure*}[t]
  \centering
    \includegraphics[width=1.0\linewidth]{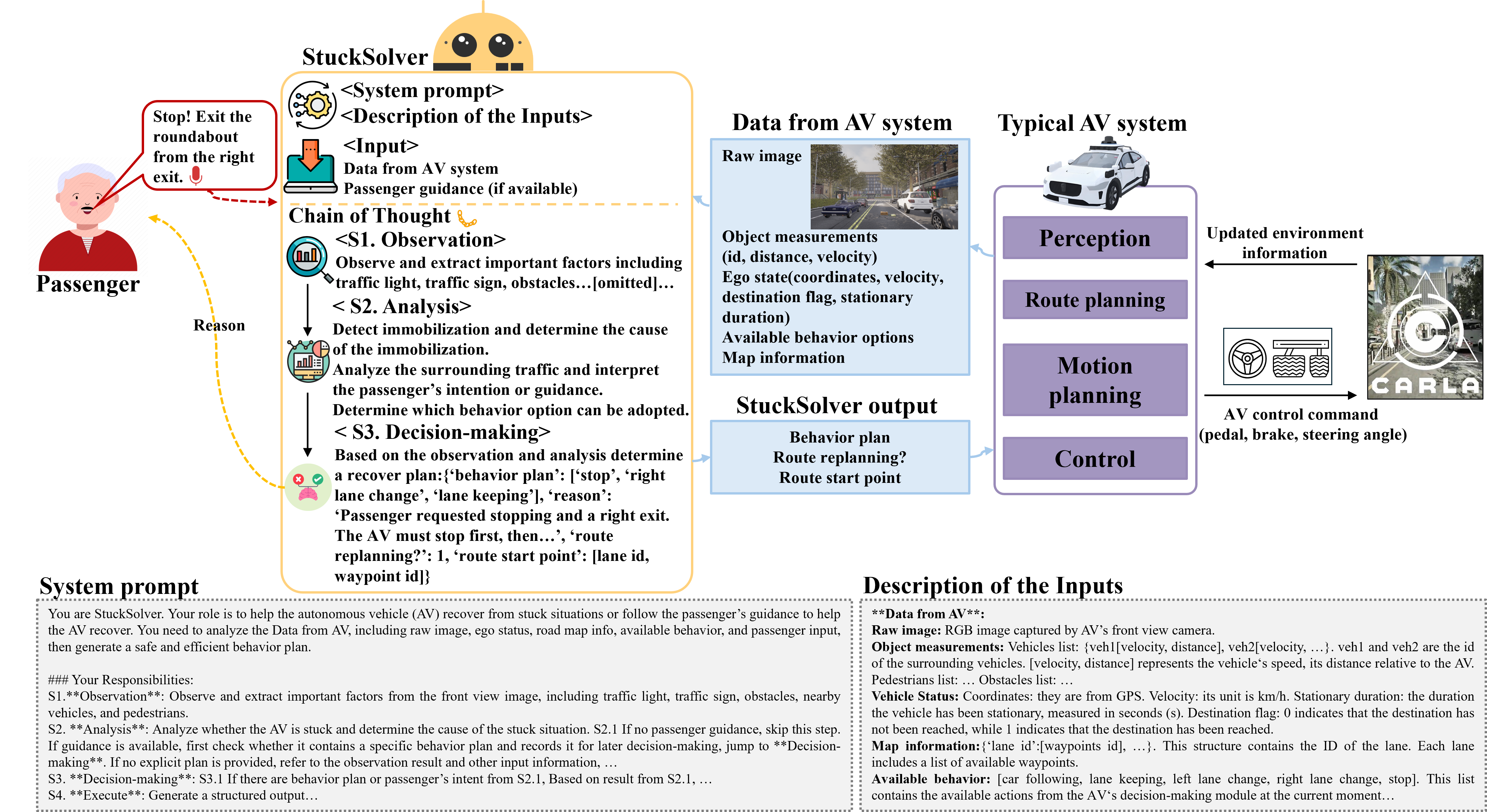}

   \caption{An overview of the LLM-assisted AV recovery method.}
   \label{fig:2}
\end{figure*}

Although AVs can complete most trips, they still face significant challenges in navigating certain traffic scenarios. Factors such as insufficient semantic understanding of surrounding entities and dense interactions at critical spatial locations can cause the decision-making or planning module to fail to generate, or mistakenly filter out, valid solutions, ultimately resulting in immobilization (i.e., becoming stationary or unable to operate normally) ~\cite{bolte2019towards,bogdoll2021description}. More critically, when this occurs, they frequently lack effective recovery mechanisms to resume normal operations, as illustrated in Fig. 1. Such incidents can cause substantial traffic disruptions and inconvenience for passengers. For example, a Waymo robotaxi in California became trapped in a roundabout after failing to generate a feasible exit trajectory. The vehicle circled 17 times before being remotely rescued by engineers ~\cite{WaymoRoundabout2024}. Passenger intervention is another recovery method. For instance, Tesla's Full Self-Driving (FSD) system exits autonomous mode upon detecting operational limits and prompts the driver to take control~\cite{TeslaAutopilot2024}. However, these existing methods have notable limitations. A remote control requires significant financial and human resources to ensure prompt assistance. Passenger intervention, on the other hand, relies on the occupant's ability to drive. This leaves elderly, disabled, or non-driving passengers without a viable solution. These challenges underscore the need for more robust and inclusive approaches to enhance AV resilience and accessibility. 

LLMs offer a promising solution due to their extensive knowledge and advanced reasoning capabilities~\cite{zhao2024large}. On one hand, LLMs have the potential to understand complex traffic scenarios and generate logical driving decisions to rescue immobilized AVs, eliminating the need for costly remote control resources~\cite{wen2023dilu}. On the other hand, LLMs can process natural language, enabling them to act as a bridge for passenger intervention~\cite{cui2024drive}. By interpreting human commands, LLMs can advise appropriate AV behaviors, making AV rides more accessible and equitable. This capability is particularly beneficial for individuals who do not or cannot drive, ensuring that AV technology serves a broader and more inclusive population. 

In this paper, we propose StuckSolver, an LLM-based recovery method designed to address AV immobilization in traffic. StuckSolver is developed as an add-on module that can be seamlessly integrated into existing rule-based AV systems, requiring minimal adaptation. It monitors the surrounding traffic conditions and the AV's operational status. When immobilization is detected, StuckSolver intervenes by generating recovery decisions and overriding the AV's original operational plans. In scenarios where StuckSolver is unable to resolve the immobilization autonomously, passengers can provide guidance, with StuckSolver acting as an intermediary to interpret and execute human commands. To evaluate its efficacy, StuckSolver is evaluated on the corner-case-rich benchmark Bench2Drive ~\cite{jia2024bench2drive}. Simulation results demonstrate that StuckSolver effectively recovers immobilized AVs, either by leveraging its own knowledge and reasoning capabilities or by interpreting passenger instructions. These findings highlight the potential of StuckSolver to enhance AV resilience in complex traffic scenarios.

\section{Related work}
LLMs have recently garnered significant attention for their remarkable potential in emulating human-like intelligence~\cite{cui2024survey}. Researchers are increasingly exploring their applications in the AV domain~\cite{wu2023language,xu2024drivegpt4,wang2023empowering,mao2023gpt,sha2023languagempc}. For instance, PromptTrack~\cite{wu2023language} integrates cross-modal features using a prompt reasoning branch to enhance 3D object tracking and prediction in driving scenes. By leveraging language prompts as semantic cues, it combines the strengths of LLMs with multi-object tracking and 3D detection tasks. DiLu~\cite{wen2023dilu} combines LLMs with a reasoning and reflection module, enabling AVs to make decisions based on common-sense knowledge. Unlike traditional learning-based methods, DiLu exhibits strong generalization capabilities by accumulating driving experience. LLMs also show promise in customizing control parameters for AV motions~\cite{cui2024board,sha2023languagempc}. For example, Sha et al.~\cite{sha2023languagempc} integrate LLMs with Model Predictive Control (MPC) to achieve adaptive motion control. Their system uses LLMs to interpret driving scenes, make decisions, and convert these decisions into actionable driving commands through a parameter matrix adaptation process. Additionally, Cui et al.~\cite{cui2024board} deploy a fine-tuned, lightweight vision-language model (VLM) on an AV, validating its effectiveness in real-world conditions. Their system adjusts control parameters based on passenger feedback after each trip, offering a personalized riding experience. These advancements highlight the transformative potential of LLMs in enhancing AV capabilities, from perception and decision-making to motion control and user interaction. 

However, most existing research focuses on isolated tasks (such as perception, decision-making, or planning), letting LLMs operate independently and evaluating their performance within only a single module. While this can demonstrate improvements to specific components of the AV system, it offers little clear evidence of the enhancement in overall driving capability. Some studies have integrated LLMs into a complete AV system~\cite{wang2023drivemlm,mao2023language} and evaluated the driving performance of LLM-powered AVs. However, those efforts primarily focus on normal driving conditions and largely neglect edge cases. Consequently, AVs may struggle in complex or unforeseen scenarios, increasing the risk of prolonged immobilization and compromising both safety and efficiency. Moreover, most of the proposed methods rarely allow passenger intervention, such as providing necessary information or decision guidance. This lack of communication misses the opportunity to further enhance the AV’s ability to navigate complex traffic scenarios.

\section{LLM-powered Recovery Method}
In this section, we present an LLM-powered recovery method designed to enable AVs to autonomously escape from immobilized situations or recover under passenger guidance through natural language commands, eliminating the need for direct driving intervention. The overall framework is illustrated in Fig. 2. 
\subsection{StuckSolver}
Our approach transforms GPT-4o~\cite{achiam2023gpt} into an intelligent agent called StuckSolver by leveraging prompt engineering~\cite{white2023prompt}, Chain-of-Thought (CoT) reasoning~\cite{wei2022chain}, and OpenAI’s Function Calling API~\cite{openai2023functioncalling}. Operating in a zero-shot mode without any task-specific fine-tuning or additional training, StuckSolver is integrated into the AV system as an add-on module. Its primary role, defined in the system prompt, is to help the AV recover from immobilization or follow passenger guidance to facilitate recovery. When immobilization is detected, StuckSolver initiates a structured CoT reasoning process that sequentially performs environment observation, multi-step analysis, and decision-making—ultimately generating a recovery plan that includes a route replanning flag, route start point, and a detailed action plan to assist the vehicle in regaining normal operation, as illustrated in Fig. 3.

\begin{figure}[t]
  \centering
    \includegraphics[width=8cm, height=12cm]{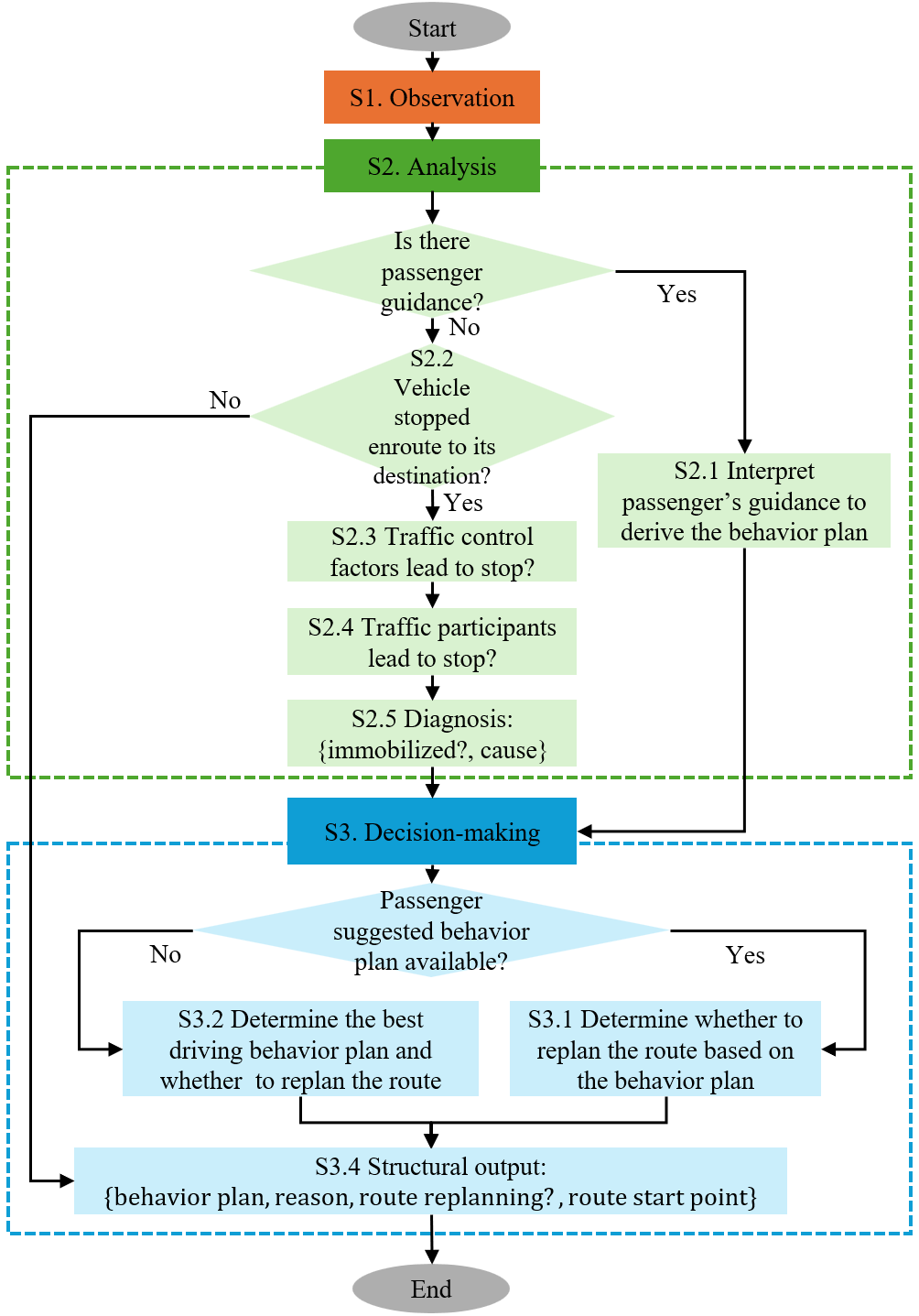}

   \caption{Chain of Thought for StuckSolver. S is an abbreviation for Step. \{immobilized? reason\} represents whether the AV is immobilized and the corresponding reason.}
   \label{fig:3}
\end{figure}

\begin{table*}[t]
\caption{Results on Bench2Drive Closed-loop Simulations}
\begin{center}
\begin{tabular}{|l|c|c|c|c|c|c|}
\hline
\textbf{Method}  & \textbf{Scheme} & \textbf{Modality} &
\multicolumn{4}{|c|}{\textbf{Closed-loop Metrics}} \\
\cline{4-7}
  &  &  & \textbf{DS $\uparrow$} & \textbf{SR(\%) $\uparrow$} & \textbf{Efficiency $\uparrow$} & \textbf{Comfort $\uparrow$} \\
\hline
TCP-traj$^{*}$ & Imitation learning & Image & 59.90 & 30.00 & 76.54 & 18.08 \\
AD-MLP  & Imitation learning & State & 18.05 & 0.00 & 48.45 & 22.63 \\
UniAD-Base & Imitation learning & Image & 45.81 & 16.36 & 129.21 & 43.58 \\
VAD  & Imitation learning & Image & 42.35 & 15.00 & 157.94 & 46.01 \\
ThinkTwice$^{*}$  & Imitation learning & Image & 62.44 & 31.23 & 69.33 & 16.22 \\
DriveAdapter  & Imitation learning & Image & 64.23 & 30.22 & 70.22 & 16.01 \\
GenAD  & Imitation learning & Image & 44.81 & 15.90 & -- & -- \\
DriveTrans & Imitation learning & Image & 63.46 & 35.01 & 100.64 & 20.78 \\
MomAD & Imitation learning & Image & 44.54 & 16.71 & 170.21 & \textbf{48.63} \\
Raw2Drive (SOTA) & Reinforcement learning & Image & \textbf{71.36} & \textbf{50.24} & \textbf{214.17} & 22.42 \\
Behavior Agent & Rule-based  & Image, LiDAR & 48.70 & 18.21 & 68.37 & 43.68 \\
Behavior Agent + StuckSolver (Ours) & Rule-based+Zero-shot & Image, LiDAR & \textbf{65.23} & \textbf{36.32} & 87.49 &  \textbf{46.74} \\
Behavior Agent + StuckSolver$^{\dagger}$ (Ours)  & Rule-based+Zero-shot & Image, LiDAR & \textbf{70.89} & \textbf{50.01} & 87.78 &  \textbf{47.31} \\
\hline
\multicolumn{7}{l}{$^{*}$ denotes expert feature distillation. $^{\dagger}$ indicates the use of human guidance in certain scenarios.} \\
\multicolumn{7}{l}{The Behavior Agent in CARLA does not include a perception module, which we add by integrating camera and LiDAR sensors.} \\

\multicolumn{7}{l}{The results of the compared methods are taken from the Raw2Drive paper ~\cite{yang2025raw2drive}.} \\

\hline
\end{tabular}
\label{tab:bench2drive}
\end{center}
\end{table*}

\textbf{Observation}. 
Observation is the first and most fundamental step in StuckSolver’s reasoning process. A thorough observation phase ensures that the system collects all necessary information before advancing to analysis and decision-making. StuckSolver captures raw images from the vehicle’s front-view camera and extracts semantic information, including traffic control factors and traffic participants. Traffic control factors ($TC$), such as traffic lights state ($TC_{tl}$), traffic signs content ($TC_{ts}$), and work zones ($TC_{wz}$). Traffic participants ($TP$), including vehicles ($TP_{v}$), pedestrians ($TP_{p}$), and obstacles ($TP_{o}$). For $TP_{v}$ and $TP_{p}$, they include type, lane position (ego lane or adjacent lane), and intent. In addition to semantic interpretation, StuckSolver augments semantic insights with quantitative measurements (e.g., distance, velocity) derived from the perception module, enabling numerically grounded reasoning and subsequent analysis.

\textbf{Analysis}.
At this stage, the goal is to determine whether the vehicle is stuck and, if so, identify the cause. The analysis process follows two possible branches: with or without passenger guidance. When passenger guidance is available (S2.1), StuckSolver interprets the passenger’s intent based on observations and formulates a behavior plan to assist the immobilized vehicle in resuming movement. In the absence of passenger guidance (S2.2-2.5), StuckSolver first checks the destination flag to determine whether it has stopped en route to its destination. If the vehicle is still in motion, StuckSolver outputs None, allowing autonomous operation to continue without intervention. However, if the vehicle has not yet reached its destination and its speed remains below the minimum threshold $v_{min}$ (set to 5km/h in this study) for longer than the stationary duration $\triangle t$ (set to 1 second in this study), it is flagged as potentially immobilized, triggering further analysis. StuckSolver begins by assessing whether any traffic control elements are influencing the vehicle’s movement. It checks $TC$ to identify possible causes of immobilization. Also, StuckSolver evaluates, based on $TP$,  whether the AV has stopped to yield to surrounding traffic participants, or to avoid a potential collision with obstacles. Finally, it provides the analysis results—indicating whether the vehicle is immobilized and the underlying cause—to the decision-making process.   

\textbf{Decision-making}.
If a passenger-suggested behavior plan is available (S3.1), StuckSolver evaluates whether route replanning is necessary to reach the destination. In the absence of passenger guidance (S3.2), StuckSolver assesses the ego state, map information, and available behavior options to generate a safe and traffic-compliant behavior plan. It also determines whether route replanning is required and identifies the appropriate starting point for the new route. Finally, the structured output of StuckSolver includes the behavior plan, the rationale behind the decision, a route replanning flag, and the designated route starting point (S3.3). 
\subsection{Integration with Autonomous Vehicle System}
We design StuckSolver as an add-on module that seamlessly integrates into existing AV systems to mitigate the inference latency challenges of LLMs. Implemented as a high-level reasoning module, StuckSolver communicates with the AV’s primary modules via structured APIs. It remains non-intrusive to the AV’s default behavior, intervening only when a stuck situation is detected. Its integration is demonstrated using the Behavior Agent in CARLA ~\cite{Dosovitskiy17}. The data flow between StuckSolver and the AV system is as follows:

\textbf{Data Flow from the AV System to StuckSolver}. The original Behavior Agent has no perception module, so we add a camera, LiDAR, Inertial Measurement Unit (IMU), and Global Navigation Satellite System (GNSS) sensors to capture surrounding objects and ego state data. The vehicle perception module directly transmits the raw image to StuckSolver, enabling it to analyze the environment ahead and extract key information. For the object measurements, they are obtained from the LiDAR with the CenterPoint algorithm ~\cite{yin2021center}. Before being passed to StuckSolver, both the object measurements and the ego state data are converted into string format and then provided as input.

In addition, the route planning module provides StuckSolver with map information, such as the IDs of the current and adjacent lanes, as well as the candidate waypoints available for route replanning. The motion planning module, on the other hand, outputs a list of currently available driving behaviors (e.g., car following, lane keeping, and lane changing ) that StuckSolver can consider when generating its recovery plan.

\begin{figure*}[t]
  \centering
   \includegraphics[width=0.7\linewidth]{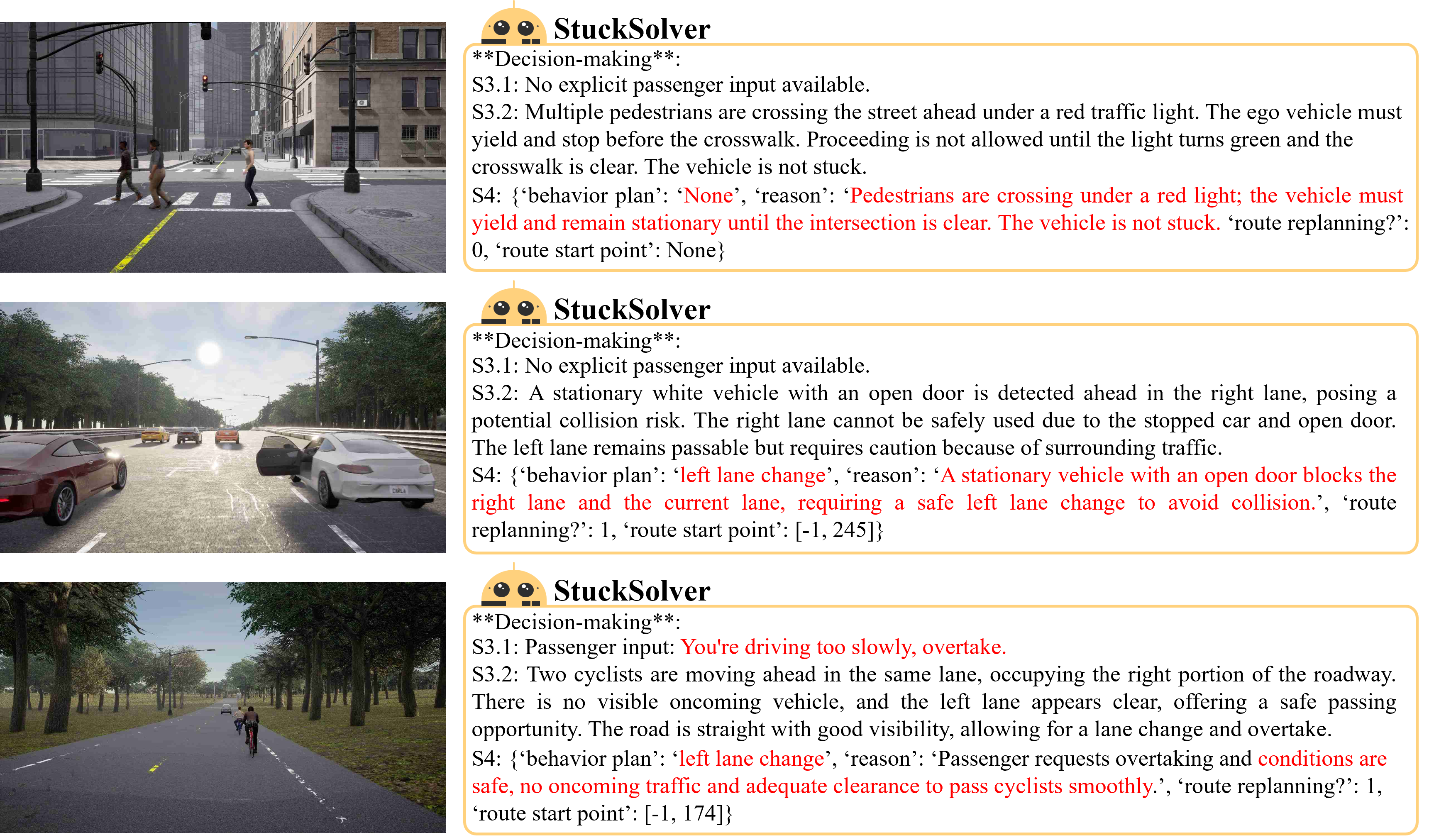}

   \caption{Experimental results for three representative scenarios: pedestrian crossing, unexpected door opening, and cyclist following. The left panels display RGB images captured by the ego vehicle’s front-view camera, while the right panels illustrate the corresponding decision-making processes of the StuckSolver.}
   \label{fig:5}
\end{figure*}

\textbf{Data Flow from StuckSolver to the AV System}. After decision-making, if the vehicle is not stuck, StuckSolver simply returns None, which will not affect the AV's original decision. If the AV is stuck, StuckSolver outputs the behavior plan as a list, replacing the AV’s original decision output. All behaviors included in the behavior plan are then sequentially executed by the corresponding control algorithms within the AV’s control module. When route replanning is necessary, StuckSolver determines suitable waypoints on specific lanes as potential starting points. It will output a replanning flag, route replanning?, and a route starting point to the route planning module for execution. 

\section{Experiment}
In this section, we conduct closed-loop driving performance evaluations using the Bench2Drive benchmark~\cite{jia2024bench2drive} within the CARLA simulator~\cite{Dosovitskiy17} (version 0.9.15). The experiments aim to quantitatively assess the effectiveness of StuckSolver in recovering immobilized autonomous vehicles. StuckSolver is built on GPT-4o-2024-08-06. All evaluations are performed on a workstation running Ubuntu 22.04, equipped with an AMD Ryzen Threadripper 7960X (24 cores) CPU and two NVIDIA RTX 5000 Ada GPUs.

\textbf{Benchmark and Metrics}.
The Bench2Drive benchmark~\cite{jia2024bench2drive} under CARLA Leaderboard 2.0 consists of 220 short routes, each containing one challenging corner case designed to evaluate different driving abilities. We adopt the official Bench2Drive metrics for evaluation. Infraction Score (IS) measures the number of violations committed along a route, including interactions with pedestrians, vehicles, road layouts, and red lights. Route Completion (RC) represents the percentage of the route completed by the autonomous agent. Driving Score (DS) is calculated as the product of Route Completion and Infraction Score. Success Rate (SR) measures the proportion of routes completed within the allotted time without traffic violations. Efficiency evaluates whether the AV’s speed is excessively low by comparing it with surrounding vehicles. Comfort assesses motion smoothness by checking whether dynamic variables such as acceleration, yaw rate, and jerk remain within expert-defined comfort thresholds that represent human driving behavior.

\textbf{Results}. The simulation results in Table 1 show that our method, which integrates StuckSolver with the CARLA rule-based Behavior Agent, achieves near state-of-the-art performance in DS (65.23), SR (36.32), and Comfort (46.74). When combined with passenger guidance, the performance further improves to DS 70.89, SR 50.01, and Comfort 47.31, demonstrating the adaptability and robustness of the proposed framework. During simulation, we continuously monitor the AV’s behavior and provide passenger instructions when we consider them beneficial, such as suggesting a lane change to navigate around an obstacle. These human-in-the-loop interactions occur in about 15\% of the 220 routes. The provided passenger guidance is not necessarily optimal and represents just one possible form of human input. Our intent is not to optimize human intervention but to demonstrate that incorporating human guidance, even in a simple and limited form, can effectively improve AV decision-making and overall driving performance.

In terms of efficiency, the Behavior Agent with StuckSolver performs below some learning based baselines(e.g., VAD, UniAD-Base, and MomAD). This difference primarily arises from the simplicity of the CARLA rule-based agent, which lacks continuous optimization and relies on discrete heuristic decision rules. Moreover, StuckSolver is activated only when immobilization risks are detected, unlike end-to-end foundation models that operate continuously and incur high computational costs. As a result, the rule-based agent may occasionally execute suboptimal maneuvers, such as missing opportunities to change to a faster lane to catch up in speed.

Overall, the results highlight a fundamental tradeoff between model complexity and operational efficiency. End-to-end approaches achieve strong performance but require substantial computation, whereas lightweight rule-based methods are efficient but less adaptive. StuckSolver bridges these two ends of the spectrum by enhancing simple and interpretable driving agents without significantly increasing computational burden. By adjusting its activation frequency through parameters such as the minimum speed threshold $v_{min}$ and the time interval $\triangle t$, the framework can shift between these extremes. A higher activation frequency makes the system behave more like an end-to-end model with the LLM engaged almost continuously, while a lower frequency allows the rule-based agent to dominate. This tunable design provides a practical means to balance computational effort and solution quality for efficient and adaptive autonomous driving.




To evaluate the rationality of StuckSolver’s decision-making process, three representative scenarios from the Bench2Drive training dataset were analyzed. In the pedestrian crossing scenario, StuckSolver correctly identifies that the ego vehicle has stopped for a red traffic light and pedestrians. It determines that the vehicle is not immobilized and appropriately refrains from intervention, producing no behavior plan. In the unexpected door-opening scenario, a stationary white sedan in the adjacent right lane has its left door fully open. After observation, StuckSolver recognizes that the ego vehicle is indeed stuck because the open door blocks its current lane. It then reasonably decides to execute a left lane change to avoid the obstruction. In the cyclist following scenario, the autonomous vehicle is moving slowly behind two cyclists. Upon receiving a verbal instruction from the passenger to overtake, StuckSolver evaluates the surrounding traffic conditions, deems the maneuver safe, and generates a left lane change behavior plan that enables the vehicle to pass the cyclists smoothly and safely.

\section{Conclusion and Future Work}
In this paper, we present StuckSolver, an LLM-based recovery framework designed to enhance AV autonomy by enabling self-recovery from immobilized situations. Unlike traditional recovery methods that depend on remote intervention or manual takeover, StuckSolver leverages the reasoning capability of LLMs to detect and resolve immobilization scenarios and interpret passenger guidance when autonomous recovery alone is insufficient. Designed as a seamlessly integrable add-on module, StuckSolver connects to existing AV systems through structured APIs, reading and writing data without modifying any internal modules or control logic. We evaluate its effectiveness on the Bench2Drive benchmark by integrating it with the rule-based Behavior Agent in CARLA. Results show that, even without passenger guidance, StuckSolver achieves DS and SR performance close to state-of-the-art end-to-end models, and that its performance further improves when passenger guidance is incorporated in specific scenarios.

Looking ahead, future work will focus on enhancing the inference efficiency of StuckSolver. Currently, its average inference time is about 2.8 seconds per query, which limits its ability to respond to time-sensitive scenarios. To address this, we plan to develop a distilled version of the underlying LLM that retains strong reasoning capability while significantly reducing latency. Another important direction is real-world testing and deployment under edge computing environments to evaluate StuckSolver’s robustness, responsiveness, and scalability in practical AV operations. In addition, we plan to further explore adaptive interaction strategies that enable StuckSolver to better interpret and learn from human guidance over time, improving its contextual understanding and decision-making in complex driving environments.


\end{document}